\documentclass[journal]{IEEEtran}
\usepackage{algorithm}
\usepackage{algorithmic}
\usepackage{xcolor}         
\usepackage{listings}

\usepackage{siunitx}
\usepackage{booktabs} 
\usepackage{multirow}
\usepackage{amsmath,amsfonts,bm}
\usepackage{color}
\lstdefinestyle{mypython}{
  language=python,
  breaklines=true,
  stepnumber=1,
  numbers=left,
  basicstyle=\fontsize{7.5}{11}\selectfont\ttfamily,
  keywordstyle=\bfseries\color{green!40!black},
}

\DeclareMathOperator*{\expectation}{\mathbb{E}}

\usepackage{cite}

\usepackage[pdftex]{graphicx}
\ifCLASSINFOpdf

\else
\fi
\usepackage{array}

\ifCLASSOPTIONcompsoc
  \usepackage[caption=false,font=normalsize,labelfont=sf,textfont=sf]{subfig}
\else
  \usepackage[caption=false,font=footnotesize]{subfig}
\fi
\usepackage{url}

\begin{document}


%
\title{Better Modelling Out-of-Distribution Regression \\on Distributed Acoustic Sensor Data \\Using Anchored Hidden State Mixup}
%
%
%

\author{Hasan Asy'ari Arief, Peter James Thomas, and Tomasz Wiktorski
\thanks{This work was funded by the Research Council of Norway’s (RCN) Petromaks2 programme (Grant number 308840) and industry partners, namely Equinor and Lundin. The computations and model reproductions were performed on resources provided by UNINETT Sigma2---the National Infrastructure for High Performance Computing and Data Storage in Norway. \textit{(Corresponding author: Hasan Asy'ari Arief.)}}
\thanks{H. A. A. and P. J. T. are with NORCE Norwegian Research Centre AS, 5008 Bergen, Norway (e-mails: hasv@norceresearch.no and peth@norceresearch.no). }
\thanks{T. W. is with the Department of Electrical Engineering and Computer Science, University of Stavanger, 4036 Stavanger, Norway (e-mail: tomasz.wiktorski@uis.no).}}

%
%

\markboth{AAM version}%
{Hidden for double-blind \MakeLowercase{\textit{et al.}}: Bare Demo of IEEEtran.cls for IEEE Journals}
%



\maketitle

\begin{abstract}
Generalizing the application of machine learning models to situations where the statistical distribution of training and test data are different has been a complex problem. 
Our contributions in this paper are threefold: (1)~we introduce an anchored-based Out of Distribution (OOD) Regression Mixup algorithm, leveraging manifold hidden state mixup and observation similarities to form a novel regularization penalty, (2)~we provide a first of its kind, high resolution Distributed Acoustic Sensor (DAS) dataset that is suitable for testing OOD regression modelling, allowing other researchers to benchmark progress in this area, and (3)~we demonstrate with an extensive evaluation the generalization performance of the proposed method against existing approaches, then show that our method achieves state-of-the-art performance. Lastly, we also demonstrate a wider applicability of the proposed method by exhibiting improved generalization performances on other types of regression datasets, including Udacity and Rotation-MNIST datasets.
\end{abstract}


%
\IEEEpeerreviewmaketitle

\section{Introduction}
\IEEEPARstart{T}{he} capability of a machine learning system to accurately model and predict data corresponding to situations with reduced similarity to those  covered by the training set is a desirable property and allows for a more reliable and safe deployment in real-world applications. However, deep neural network as the backbone of state-of-the-art machine learning systems often provide incorrect predictions but report falsely high confidence when evaluated on distributional shifts, also called Out-of-Distribution (OOD), dataset~\cite{likelihood-ratios-google}. This is problematic because the distribution of the real-world data often covers a much wider range of characteristics compared to those covered by carefully curated training datasets. Therefore erroneous and high confidence predictions are a major roadblock when implementing machine learning techniques in applications sensitive to safety, security and cost.

\begin{figure}[t]
    \centering
    \includegraphics[width=0.99\columnwidth]{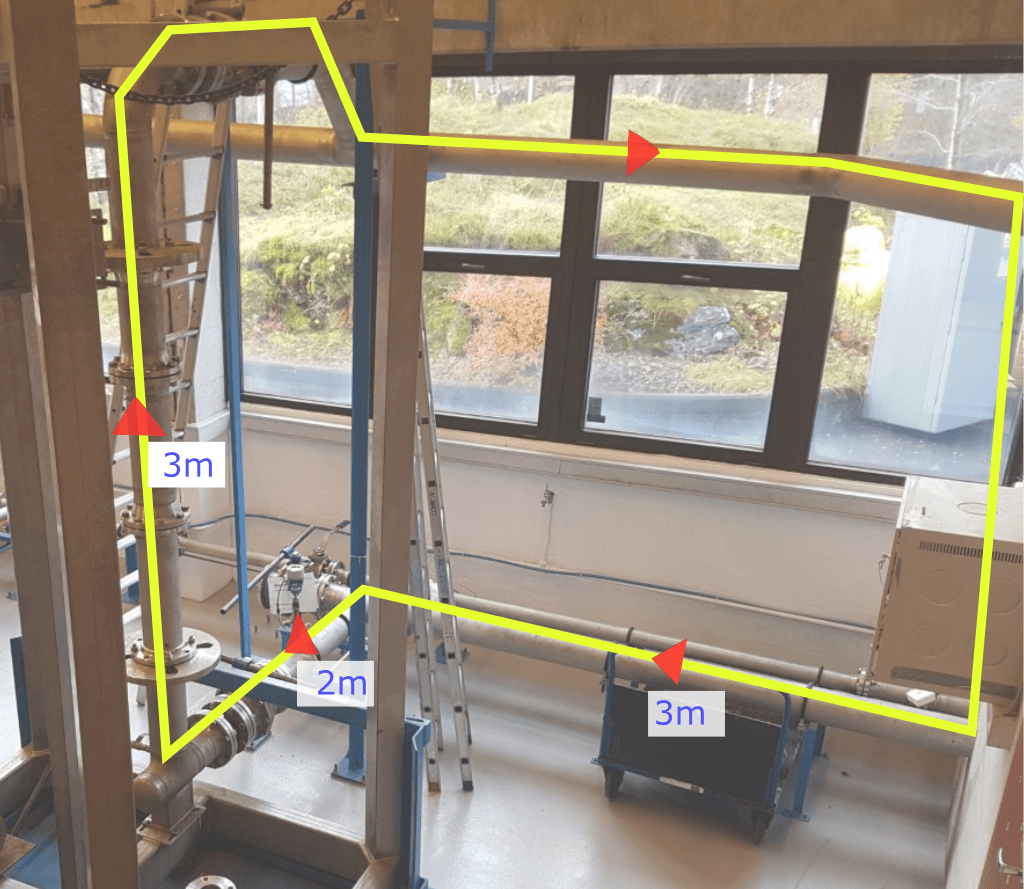}
    \caption{Fiber optic cable (in yellow) installed on the multiphase flow loop infrastructure, used for collecting the MPFF-DAS dataset.}
    \label{fig:fiber-optic-installment}
\end{figure}
\begin{figure}[t]
    \centering
    \includegraphics[width=0.95\columnwidth]{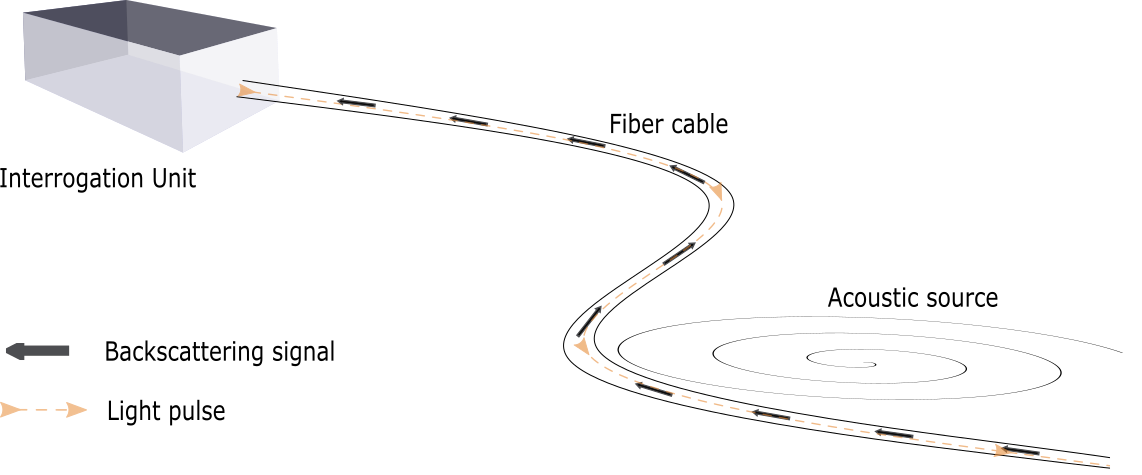}
    \caption{Illustration of a fiber optic cable as a Distributed Acoustic Sensor. The Interrogation Unit (IU) sends light pulses along the fiber, the backscattering signal travels back to the IU carrying acoustic profile along the cable, depicted from~\cite{arief2021survey}.}
    \label{fig:fiber-opticIU}
\end{figure}
\begin{figure}[t]
    \centering
    \includegraphics[width=0.95\columnwidth]{{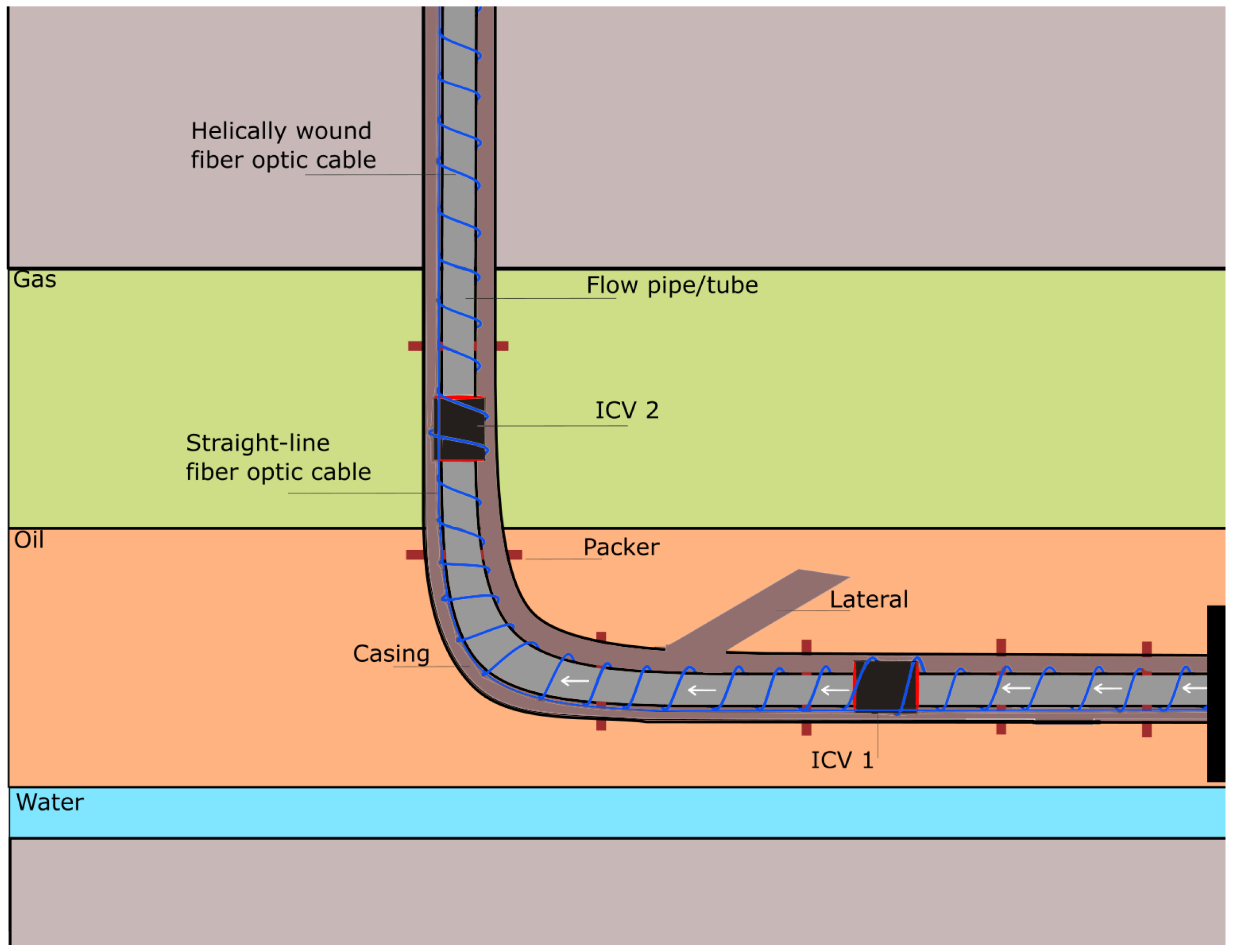}}
    \caption{Sketch of fiber optic instalment for down-hole measurement using Distributed Acoustic Sensor (DAS), depicted from~\cite{arief2021survey}.}
    \label{fig:fiber-optic-utilizations}
\end{figure}

Collecting more data from diverse scenarios in multi-environment settings is one method to provide robust generalization and can reduce the OOD problem~\cite{arjovsky2019invariant}. However in many cases, data collections are expensive and it is often impossible to capture all possible scenarios. For example, in the case of distributed phase-fraction estimation of multiphase fluids, collecting training data from across a wide range of  combinations of Water in Liquid Ratio (WLR), Gas Volume Fraction (GVF), fluid flow rate, fluid velocity, pressure and temperature setting, pipe diameter, etc., will lead to an intractable test matrix and experimental design. Fig. \ref{fig:fiber-optic-installment} depicts a fiber optic installation for collecting the acoustic signature of multiphase fluid flowing in a flow-loop infrastructure. Fig.~\ref{fig:fiber-opticIU} shows how the fiber optic cable acts as a sensor array for detecting acoustic sources within the surrounding environments. The distributed phase fraction is defined as the percentage (or fraction) of water, oil, and gas in the total mixture of fluids within the pipeline. The phase fractions are represented by the WLR and GVF. Distributed phase fraction measurements can provide a game-changing sensing capability in the multi-billion dollar hydrocarbon production industry, and can be deployed as depicted in Fig.~\ref{fig:fiber-optic-utilizations}. The technology also provides  environmental benefits such as reducing carbon footprints due to production and reducing the need for new oil field developments while renewable alternatives reach maturity~\cite{arief2021survey}.

A popular and intuitive way to address the generalization problem is by enhancing the regularization capacity of the machine learning system, for example, by using dropout~\cite{dropout}, zero-shot learning~\cite{zeroshot}, label smoothing~\cite{szegedy2016rethinking}, as well as regularization norms, such as Ridge Regularization and Lasso Regularization. 
In addition, \cite{arjovsky2019invariant} introduced Invariant Risk Minimization (IRM) and~\cite{krueger2020out} proposed Risk Extrapolation (REx) that were trained on multi-environment settings to provide robust generalization outside the training data. Recently, several novel data augmentation strategies have also been proposed, including Input Mixup~\cite{inputmixup}, Manifold Mixup~\cite{manifold-mixup}, AdaMixup~\cite{adamixup}, and Output Mixup as an activation function~\cite{outputmixup}. These papers provided strong experimental evidence for better generalization of the neural network models by interpolating within the training data as well as within the manifold hidden variables. 

\begin{figure}[t]
    \centering
    \includegraphics[width=1.0\columnwidth]{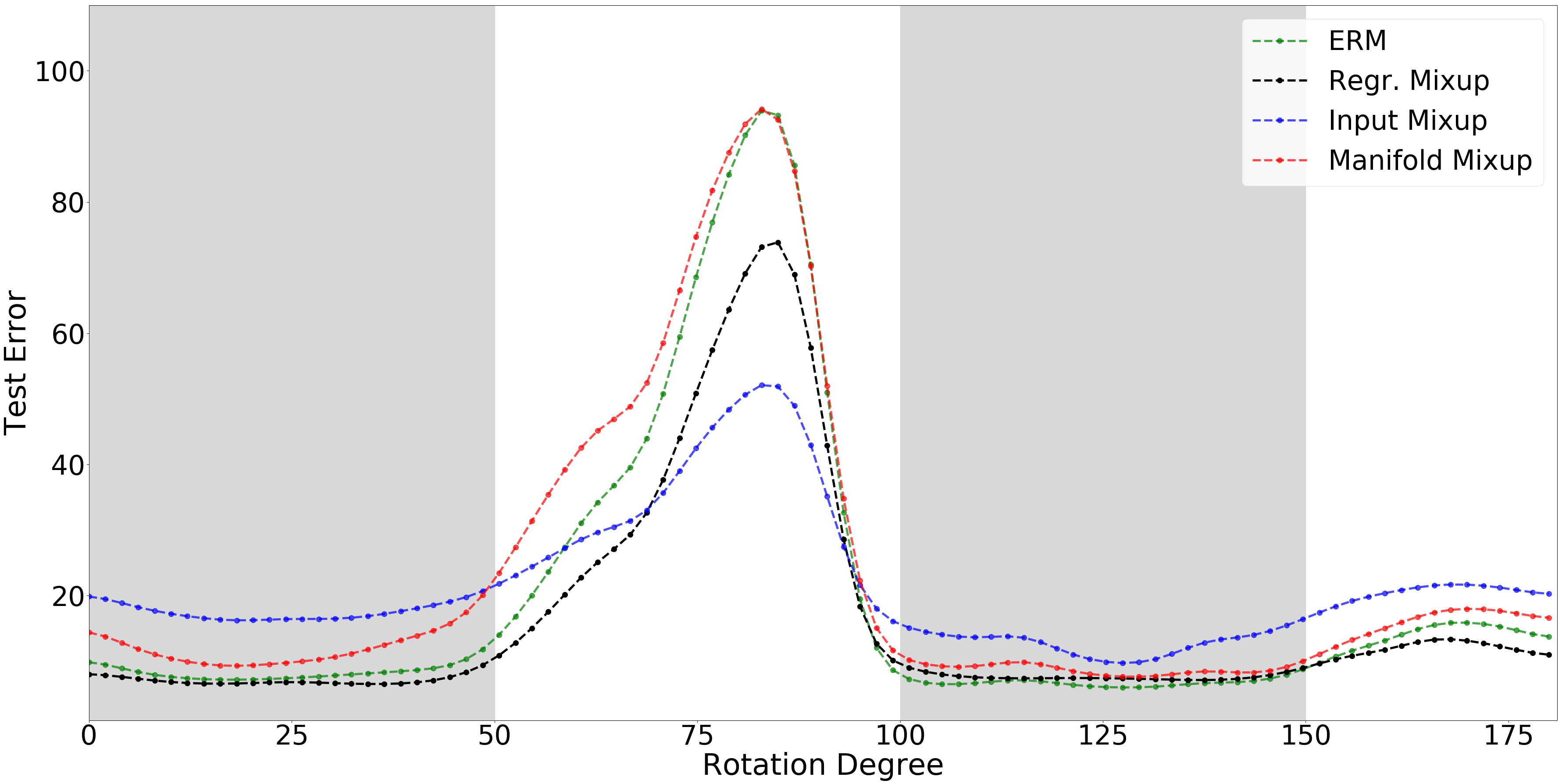}
    \caption{Distribution of the test error along the rotation degrees on the Rotation-MNIST dataset. The grey areas show the existing samples in the training data, while the white areas show the OOD test data.}
    \label{fig:test-mnist-distribution}
\end{figure}

Building on the interpolation and regularization ideas, this paper proposes a novel technique called OOD Regression Mixup, specifically developed for reducing the distributional shift problem on regression-based datasets. In the regression problem, the distance of the target from two randomly picked data points consists of a range of real values. These values are also the target variables from different data points. The mixup algorithms mostly ignore this underlying condition when interpolating between two data points while performing the augmentation. The proposed OOD Regression Mixup picks up the interpolation idea and builds on the linearity assumption from~\cite{inputmixup} to proportionally weight the manifold hidden variables of neural network using a contrast sensitive distance kernel from an anchored data point and use this as a regularization signal. We demonstrate that the method provides a strong regularization capacity on regression-based datasets, depicted in Fig. \ref{fig:test-mnist-distribution}. While our work was motivated by modelling the WLR using Distributed Acoustic Sensing (DAS), we found out that our methodology is applicable in other settings, including image regression datasets. The contributions of our work are as follows:
\begin{itemize}

\item We introduce a novel anchored-based OOD Regression Mixup algorithm that proportionally weights the manifold hidden variables from a neural network model, leveraging a distance-based kernel and providing a more generalized capability of the machine learning system.

\item We provide a unique 450~GB spatio-temporal distributed acoustic sensor dataset from multi-phase fluid flow experiments (also called MPFF-DAS). 
The dataset covers a wide range of phase-fraction situations and is highly suitable for validating generalization techniques for the complex OOD regression problem.

\item We demonstrate our proposal with an extensive evaluation on modelling the OOD datasets, including on the MPFF-DAS, Udacity and rotation-MNIST datasets. We also evaluate several novel generalization techniques on the datasets and show that our method achieves state-of-the-art performance on this challenging problem. 

\end{itemize}


\section{Background}
\begin{figure}

  \centering
  \subfloat[Distribution of WLR on the training data.\label{fig:training-distribution}]{%
       \includegraphics[width=0.9\linewidth]{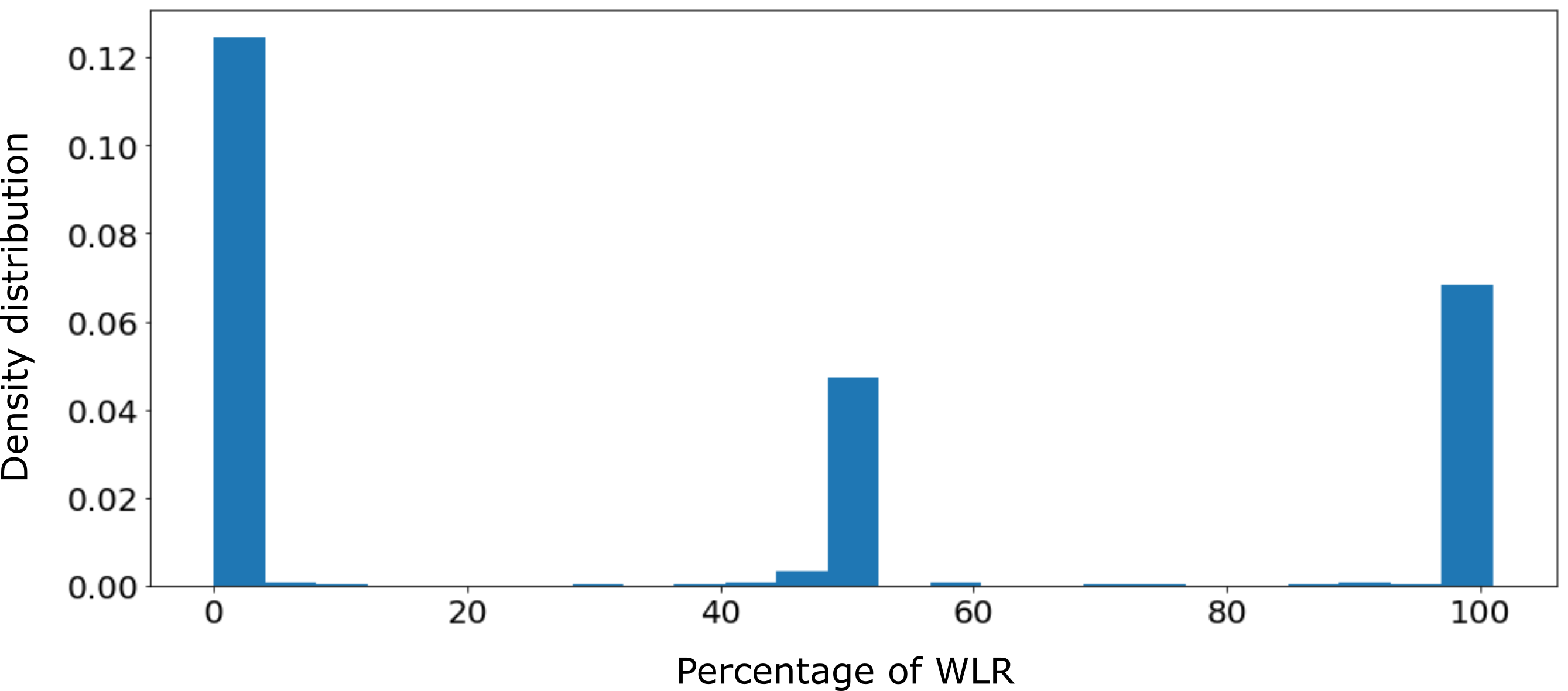}
       }
    \hfill
  \subfloat[Distribution of WLR on the test data.\label{fig:test-distribution}]{%
       \includegraphics[width=0.9\linewidth]{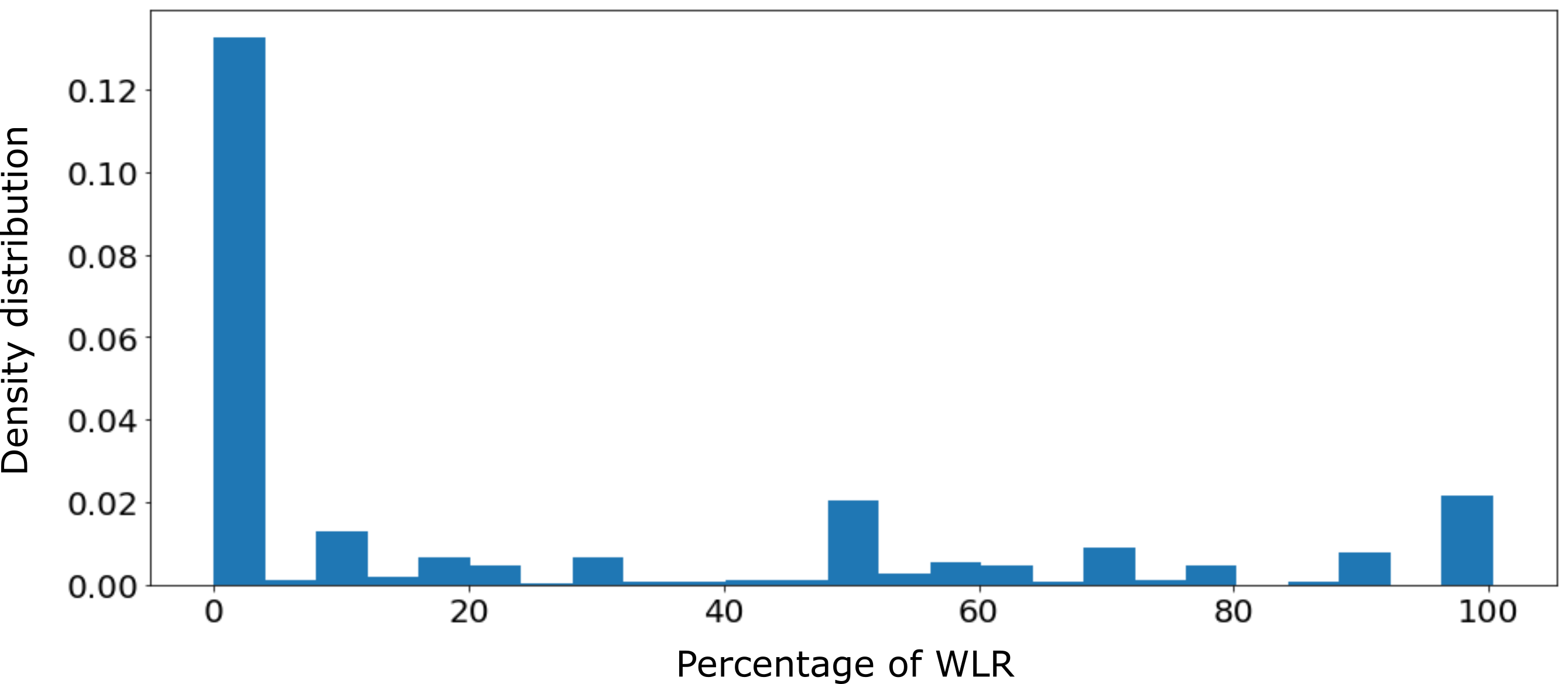}
       }
    
	\caption{The MPFF-DAS data distribution for phase-fraction information in terms of WLR. The $x$-axis represents the percentage of water in the multiphase fluids, while the $y$-axis represents the density distribution from the total data point. 
	}
	\label{fig:data-distribution}
\end{figure}

Suppose we have a training set $D$ of $(x,y)$ pairs sampled from the true distribution $(x, y) \sim P$, where $x$ is raw input vector and $y$ is a real value ranging between two real values. Modelling the $(x,y)$ pairs is a regression task, with an objective to find a minimum $p(x) - y$, similar to~\cite{susto2015supervised}.
The OOD setting is considered when the $y$ sampled from $P$ does not belong to $D$, meaning that the $p(x)$ is shifting while $p(y|x)$ is staying the same.
In this paper, the focus is mainly on the datasets that have low similarity (high distributional shift) with the test set. Fig.~\ref{fig:data-distribution} shows the difference of data distribution between the training and test data from the MPFF-DAS dataset, also called the OOD DAS dataset.

Deep neural network algorithms are trained to minimize the average error over the training data using a rule known as Empirical Risk Minimization (ERM) principle~\cite{ERM-vpnik-1998}. The ERM is used to approximate expected risk by calculating error using the  loss function $\ell$ over the true distribution $P$. The approximation is used because the distribution $P$ is usually unknown, especially in complex systems. Therefore, the ERM uses the empirical distribution from the training data to calculate the empirical risk $R_\delta(f)$, defined as:
\begin{equation}
    R_\delta(f) = \frac{1}{n}
    \sum_{i=1}^n \ell(f(x_i), y_i). \label{eq:erm}
\end{equation}
However, focusing only on fitting well to the training data and ignoring the potential that training data does not represent $P$ will lead to overfitting and memorization~\cite{memorization-neural-network}.  OOD modelling techniques, called IRM and Rex, works by using multi-environment training sets to minimize the variance of ERM from multi-set training distribution, with aim of achieving minimum variance on the new unseen distributional shift data points.

Other generalization techniques build on the assumption that the training data is not enough to represent the true distribution of the overall objective, therefore new data or slight variations of the training data can help overcome this limitation. Data augmentation strategies build on this idea by introducing a new representation of the training data. A classical augmentation technique is implemented by rotating, colour shifting, flipping, and blurring the input image~\cite{alexnet}. It can also work by introducing noise perturbation within the data~\cite{purwins2019deep}, or by combining augmentation with random field algorithm~\cite{chen2021effective}. A more recent data augmentation strategy is the mixup algorithm~\cite{inputmixup}. It is based on the Vicinal Risk Minimization (VRM) principle~\cite{chapelle2000vicinal} which uses virtual training set from vicinity distribution $(v)$ to approximate the true distribution $P$. The input mixup method samples from a vicinal mixup distribution $(x,y) \sim v$ by interpolating pairs of input data $(x_i, y_i)$ and $(x_j, y_j)$  controlled by a random variable $\alpha$ to produce a new pair of input-output representations $(\tilde{x}, \tilde{y})$. Specifically, the sampling procedure is as follows:
\begin{align*}
  \tilde{x} &= \lambda x_i + (1 - \lambda) x_j,\\
  \tilde{y} &= \lambda y_i + (1 - \lambda) y_j,
\end{align*}
where $\lambda \sim \text{Beta}(\alpha, \alpha)$, for $\alpha \in (0, \infty)$. The $\alpha$ represent the strength of interpolations on the input and output pairs. The other type of mixup is called AdaMixup which learns the mixing policy regions automatically, and benefits from mixing multiple inputs for generalization. Lastly, the manifold mixup focuses on the mixing pairs of hidden variables within the neural network layers with the aim of improving the hidden representation and decision boundaries of the neural network layers. For completeness, as the manifold mixup aims at modelling the hidden variables, several feature clustering algorithms can also be considered for optimizing the hidden variables within the feature spaces. The techniques such as, rank-constrained clustering algorithm \cite{rankconstrained} as well as  dynamic affinity graph construction\cite{dynamicaffinity}, can be considered; the latter has the ability to deal with redundant visual features which could be interpreted as part of the generalization techniques.

\section{Regression model with Mixup}

In the next section, we will show some limitations when using existing mixup algorithms for modelling regression datasets (Section \ref{mixup-limitation}). We will also show how we overcome the problem using anchored-based  OOD Regression Mixup (Section \ref{anchored-mixup}), and why using our technique is more desirable than using the other mixup algorithms for modelling the regression data~(Section \ref{desirable-property-of-AMp}).

\subsection{Rethinking Mixup \label{mixup-limitation}}
In a regression task, the smaller the distance between two target values, the more similar these target values are, unless they represent different objects.  
In contrast, in a pure classification task, the class numbers only represent the ID of each class and have no further meaning. With that in mind, when sampling from $v$ in the discretized regression or pure regression task, the virtual target $\tilde{y}$ can be an actual target value lies between $y_i$ and $y_k$. Let $\tilde{y}= y_j$ and $y_i~\leq~y_j~\leq~y_k$, then the~$y_j$ sampled from training data $D$ which is a pair of ($x_j, y_j$), should have some weight when a model is trained using the VRM principle to minimize the vicinal risk, as explained in~\cite{inputmixup}. The input mixup algorithm, however, ignores this underlying condition by not taking into account the potential contribution of $y_j$, when the model was trained on pairs of ($x_i, y_i$) and ($x_k, y_k$). The manifold mixup also ignores (during training) the empirical risk from $(x_j, y_j)$ when mixing the hidden states from the aforementioned pairs.

Most of the existing mixup ideas justify the interpolation approaches by arguing that they can provide better separation between target classes or manifold hidden variables from two different classes. None of them highlighted that the linear interpolation function in the mixup algorithms might work due to the existing linear correlation between two data points. In regression tasks, linear mapping and correlation between two data points exist to some degree and can be quantified by the spatial autocorrelation principle. These similarity and linearity assumptions are based on Tobler’s First Law of Geography: \textit{"All things are related, but nearby things are more related than distant things"}~\cite{tobler2004first}. Finally, we hypothesize that "{the neighbourhood similarities, calculated using distance kernel from $y$, can help better generalize regression model}". 

\subsection{Anchored-based  OOD Regression Mixup \label{anchored-mixup}}
\begin{figure}[t]
    \centering
    \includegraphics[width=0.9\columnwidth]{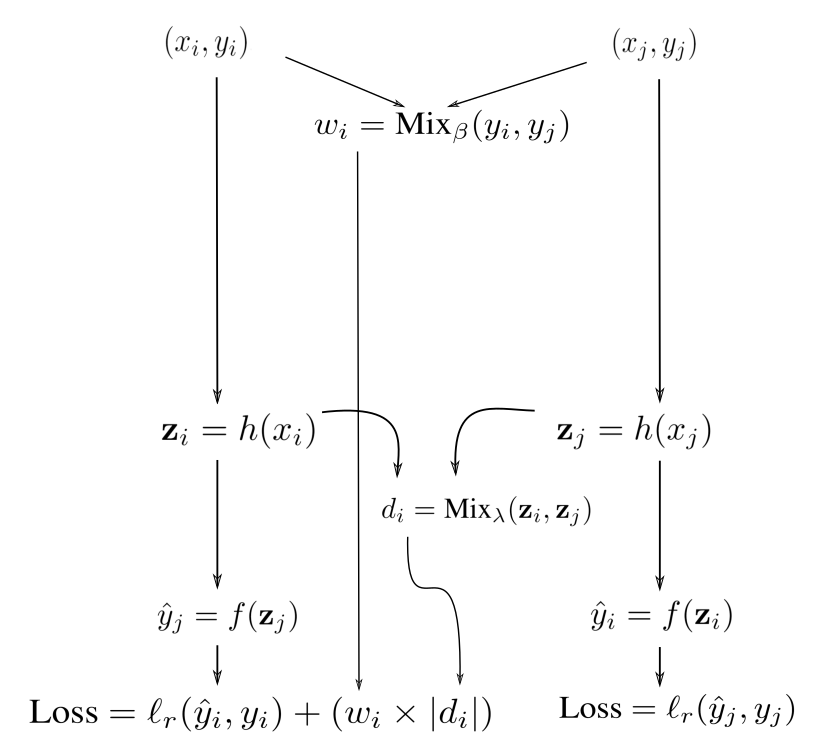}
    \caption{Schematic of OOD Regression Mixup. $\text{Mix}_{\beta}$ and $\text{Mix}_{\lambda}$ are explained in Eq. \Ref{eq:contrast_sensitive_kernel} and Eq. \Ref{eq:proportional_z}, respectively. It is worth noting that $\lambda$ for OOD Regression Mixup is not sampled from $\text{Beta}(\alpha, \alpha)$ but it is a fixed number to control the regularization penalty.}
    \label{fig:mixup:schematic-comparisons}
\end{figure}

Directly interpolating input and output pairs in regression datasets depends heavily on  the pairs having a strong linear correlation with each other, which is rarely the case in modelling the real-world data. Therefore, rather than focusing on input mixup, the proposed OOD Regression Mixup is implemented in the manifold hidden variables within the  neural network. It should be noted that the proposed algorithm does not operate as an augmentation technique, rather it behaves more like an additional cost function mixing the distance between two hidden variables from different data points as a regularization signal, depicted in Fig. \ref{fig:mixup:schematic-comparisons}. 

In this paper, the neural network is trained as 
\begin{align*}
f(h(x)) = \hat{y} ,
\end{align*}
where $h(x)$ is all the layers of neural network before the fully connected layer and $\hat{y}$ denotes the neural network prediction. The output of $h(x)$ is denoted as \textbf{z}, and it represents the $n$-dimensional feature vector before the final regression head. 

The OOD Regression Mixup minimizes the proportional distance between hidden variables $\textbf{z}_i$ and $\textbf{z}_j$ from two different data points $(x_i, y_i)$ and $(x_j, y_j)$, to enforce the linearity assumption of two similar target variables,~$y_i~\sim~y_j$. The pair $(x_j, y_j)$ is called an anchored data point, because it is used as a pseudo-target variable when calculating the proportional distance between $\textbf{z}_i$ and $\textbf{z}_j$. It must be emphasized that in OOD Regression Mixup, $(x_j, y_j)$ is an actual data point sampled from the training distribution $D$ and trained using the normal ERM principle.

The algorithm is implemented in three steps. First, we calculate the distance between $y_i$ and $y_j$ using a contrast sensitive distance kernel. It is defined as:
\begin{align}
\label{eq:contrast_sensitive_kernel}
    w_i = & exp(- \frac{|y_i - y_j|}{\beta^2} ) ,
\end{align}
where $\beta$ is a fixed number and denotes the limit of the mixup effect. The larger the distance between $y_i$ and $y_j$ the smaller the mixup effect controlled by $\beta$. The distance kernel is a simplification of the smoothness kernel from the Conditional Random Field (CRF) algorithm~\cite{arief2019addressing} and the kernel is used because we aim to invoke a similar effect of CRF on measuring contrast smoothness of the targeted (smooth) regression data.

Next, we calculate the proportional distance between $\textbf{z}$ when compared to $y$ denoted as $d$. With $\lambda$ denotes the regularization learning rate controlling the overall effect of the regularization objective, the $d_i$ is defined as:
\begin{align}
\label{eq:proportional_z}
    d_i = &  \frac{\lambda}{n} \sum_{m=0}^n \frac{y_i \times \textbf{z}_{jm}  -  y_j \times \textbf{z}_{im}}{y_j \times \textbf{z}_{jm} },
\end{align} 
with aim to achieve
\begin{align*}
    \frac{y_i}{y_j}  \sim  \frac{\textbf{z}_i}{\textbf{z}_j}, \text{ when } y_i \sim y_j.
\end{align*}
It is worth mentioning that the proportional distance is used, instead of other distance metrics e.g. L1 and L2 distances, because it ensures smooth penalty on small-range similar data points while working fairly well on tackling anomalous features on high dimensional feature vector. 

Finally, we calculate the OOD Regression Mixup as an additional cost function $( w_{i} \times |d_{i}|)$, therefore the neural network minimizes:

\begin{align}
\label{eq:regression_mixup}
    L(f) = &\expectation_{(x_i,y_i) \sim P}\,
    \expectation_{(x_j, y_j) \sim P}\,
    \ell_r(\hat{y}, y_i) + ( w_{i} \times |d_{i}|).
\end{align}

\subsection{Theoretical explanation \label{desirable-property-of-AMp}}

\begin{figure}[t]

\begin{lstlisting}[style=mypython]
# lam and beta are fixed numbers.
for (xi, yi), (xj, yj) in zip(loader_i, loader_j):
    Zi, Zj = net.last(xi), net.last(xj)
    ERM = [l(net.fc(Zi), yi), l(net.fc(Zj), yj)]
    # prevent division by zero
    if ( yj*Zj !=0 ).all(): 
        wi = exp(-((yi-yj).abs()/beta**2))
        di = ((yi*Zj - yj*Zi)/(yj*Zj)).mean() * lam
        ERM[0] = ERM[0] + (wi * di.abs())
    optimizer.zero_grad()
    ERM.mean().backward()
    optimizer.step()
\end{lstlisting}
    \caption{Mixup Regularization training procedure in PyTorch.}
    \label{fig:mixup:code}

\end{figure}

The mixup algorithms have shown numerous successes in augmenting the training data to achieve better generalization, not to mention they also improve calibration and predictive uncertainty~\cite{mixupuncertainty}. The simplistic nature and the minimum memory overhead are additional advantages of this powerful yet robust algorithm.  In classification tasks for example, mixing Car and Apple classes could improve the separation of the decision boundary between the two classes. In regression tasks however, that behaviour could have an inverse effect, especially in smooth regression datasets. The decision boundary between the two closest neighbouring targets should be minimized because they might share interdependent similarities. 

Let $\boldsymbol{z}_0$ and $\boldsymbol{z}_1$ be the component of $\textbf{z}$, in other words 
\begin{align*}
\textbf{z}~=~[z_0~z_1~...~z_{n-1}~z_n]; 
\end{align*}
$\textbf{z}$ is the discriminative feature vector extracted from the input data of a regression model. Then, there will be $\boldsymbol{z}_a \in \textbf{z}_i$ and $\boldsymbol{z}_b \in \textbf{z}_j$ where $\boldsymbol{z}_a \times w_a$ proportionally comparable to $\boldsymbol{z}_b \times w_b $  when $y_i \sim y_j$, otherwise the regression transition is not smooth.  $w_a$ and $w_b$ denote the fully connected weights in the neural network that maps $\textbf{z}$ to the prediction $\hat{y}$. By separating between two target (or classes), the Manifold Mixup unfortunately increases the distance between $\boldsymbol{z}_a$~and~$\boldsymbol{z}_b$, therefore $\boldsymbol{z}_a \times w_a \neq \boldsymbol{z}_b \times w_b $. The proposed mixup will make sure $\boldsymbol{z}_a \times w_a$ be more comparable to $\boldsymbol{z}_b \times w_b$ by targeting the proportional distance of $y_i / y_j \sim \textbf{z}_i / \textbf{z}_j$ in Eq. \ref{eq:proportional_z} when minimizing the overall objectives during training.

\begin{figure}[t]
    \centering
    \includegraphics[width=0.95\columnwidth]{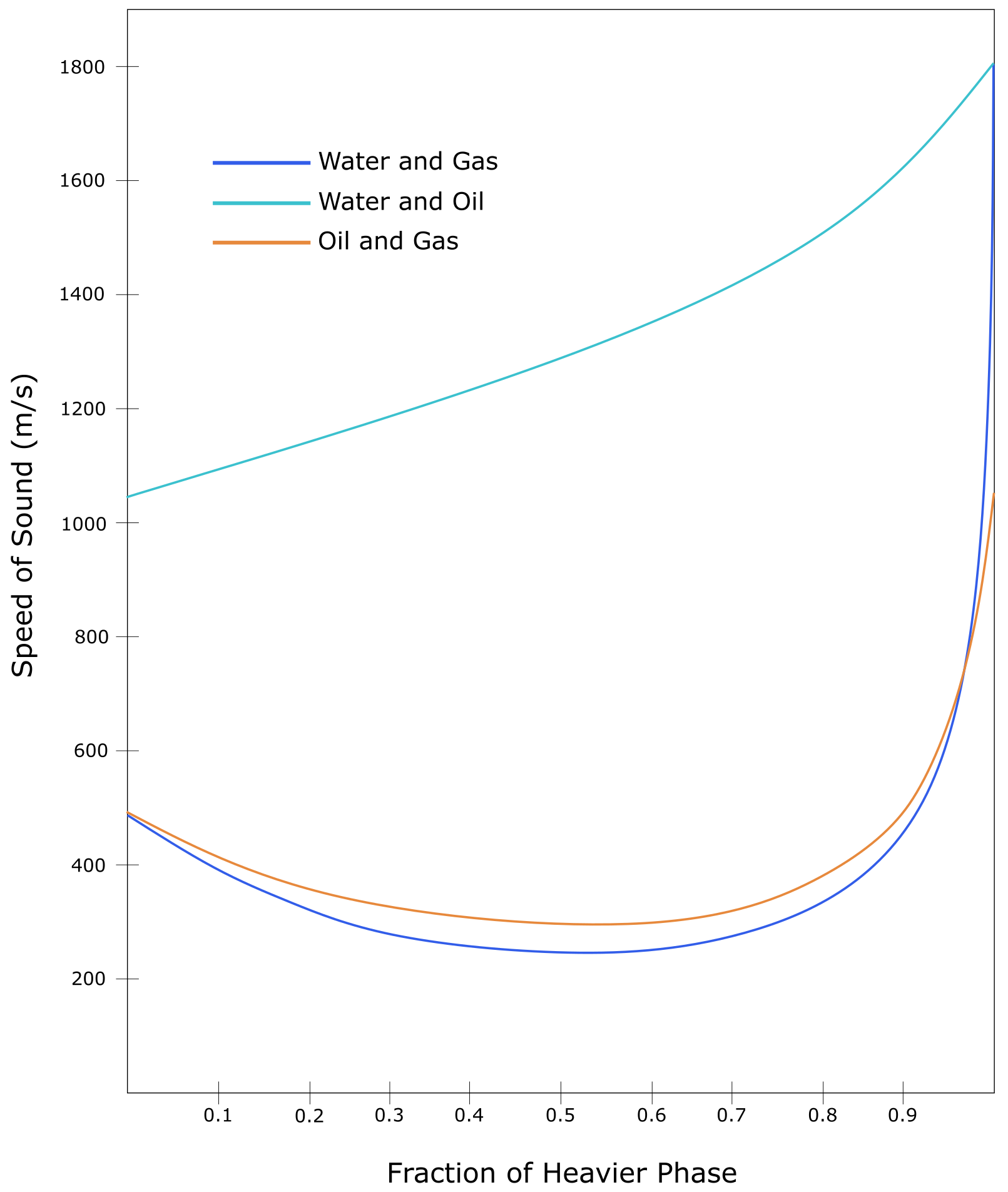}
    \caption{Relation between Speed of Sound and multiphase mixture including WLR and GVF, depicted from~\cite{arief2021survey}.}
    \label{fig:sos-maps}
\end{figure}

The $\boldsymbol{z}_a$ and $\boldsymbol{z}_b$ can be thought of as the subset of discriminative features that up to some limit can linearly map the target variables; the mapping limit in  OOD Regression Mixup is controlled by $\beta$. In modelling the WLR from DAS, for example, the Speed of Sound (SoS) can provide smooth linear mapping among the closest phase-fraction values, depicted in Fig. \ref{fig:sos-maps}. Extracting SoS, however, is not trivial and accurate SoS estimation can only be achieved within certain constraints, including having sufficiently high Signal to Noise Ratio (SNR) data covering a sufficient number of spatial channels.

Other examples are found in the Udacity and rotation-MNIST datasets. The Udacity dataset (Udacity 2018) is used for detecting the steering angle of a car from visual representation seen by the driver outside the car.\footnote{The Udacity dataset was published under MIT license and is available at https://github.com/udacity/self-driving-car.} This dataset shows a smooth transition between two closest angles when they are estimated from the same scene and location. The rotation-MNIST dataset, on the other hand, was built by rotating (between 0 and 180 degrees) MNIST images~\cite{mnist}; the MNIST dataset was published under CC BY-SA 3.0 license. While it is a toy dataset, it simulates a smooth regression phenomenon. The smaller the distance of the rotation angle between two data points, the more similar the images are, unless the two images represent different objects or class numbers which is common occurrences in multivariate regression.
While those examples do not capture all the different behaviours of the regression dataset, we hypothesize that there exists a smooth transition (a.k.a linear correlation) within a small range of neighbouring targets. We quoted from~\cite{inputmixup}: "\textit{Linearity is a good inductive bias from the perspective of Occam's razor since it is one of the simplest possible behaviours}". Therefore, taking advantage of this condition will provide a better generalization for the given modelling objectives.

As long-range linearity is often not the case for complex real-world data, we argue that the OOD Regression Mixup will be more applicable for OOD regression rather than the normal regression data. In the OOD setting, there are too many unknowns, and when it is OOD, we can safely assume that the data distribution of training and test data is different, hence the term out-of-distribution. Our proposal invokes a common property of smooth regression data: the data with similar target variables should have a similar correlation between features/properties. For example in fluid dynamics, if fluid temperature increases slowly then the fluid volume will also expand slowly. The correlation between temperature and volume in physics is obvious and can be applied to many applications. We argue that bringing this general formulation of linear correlation to the OOD data is more advantageous than methods aiming to fit a model only on training data that has a different distribution compared to the test data. In general regression modelling, the test data is relatively similar to the training data. Therefore, forcing the general formulation of linear correlation using our proposal might be problematic if the training data in actuality cannot be linearly mapped.

\section{Experiments}
To validate our hypothesis, we will turn to the empirical evidence of  the OOD Regression Mixup. In Section \Ref{das-dataset}, we introduce our unique distributed acoustic sensor dataset. In Section \Ref{experiment-wlr-das}, several data generalization techniques were employed to model the DAS data to estimate the WLR. In Section \Ref{experiment-on-image-dataset}, we show that the proposal can be used further to generalize regression models on other types of datasets. In Section \ref{computation-cost}, we discuss the computation cost from our algorithm, while in Section \ref{ablation-study} we provide ablation experiments analysing the sensitivity of parameters used in our algorithm. Lastly, in Section \ref{limitation}, we discuss some limitations when modelling regression data using the proposed mixup.

\begin{figure}[t]
    \centering
    \includegraphics[width=0.99\columnwidth]{{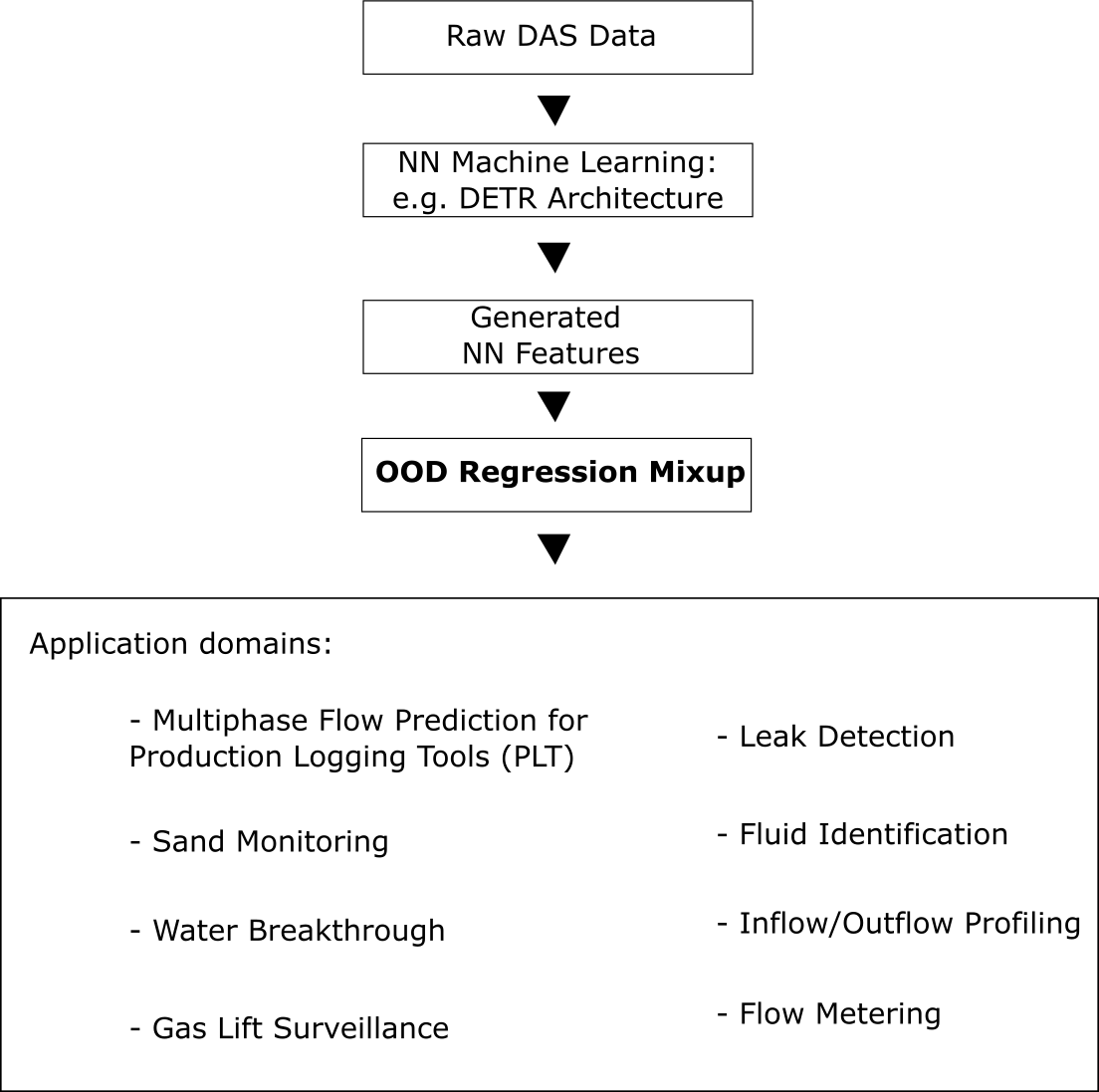}}
    \caption{Block diagram for utilization of DAS data for hydrocarbon monitoring industry using machine learning and OOD Regression Mixup.}
    \label{fig:block-diagram}
\end{figure}

\subsection{MPFF-DAS dataset}\label{das-dataset}
The Multiphase Fluid Flow - Distributed Acoustic Sensor (MPFF-DAS) dataset is high-resolution spatio-temporal data, consisting of a high-frequency temporal dimension (40khz) and high-resolution spatial sampling (0.4~m/sample, 5m gauge length). The dataset contains acoustic information on every location covered by the fiber optic cable. The dataset was obtained from controlled experiments in a flow-loop laboratory with the fiber cables attached in a straight line on the flow-loop pipes.
The training and test data were taken at different times with different experimental designs, therefore, they represent the distributional shift behaviour. The training and test data consist of two and four hours of recording, respectively. 

The dataset consists of phase-recovered DAS data and log files from the experiments consisting of multi-phase information, including WLR, GVF, fluid flow rate, pressure, temperature, timestamp, and flow velocity. The phase-recovered DAS data with their corresponding labels are in the compressed Numpy format consisting of 30 seconds recording with a median file size of 0.7~GB per file and have been sliced into 5-second intervals with 3-seconds overlap, therefore they are ready to be used for machine learning models. The total size of the MPFF-DAS dataset is around 450~GB and as far as we know, it is the only publicly available DAS dataset relevant to multi-phase fluid flow characterisation.  

\subsection{Generalization on estimating WLR using DAS}\label{experiment-wlr-das}

In this paper, the MPFF-DAS dataset was used to estimate the distributed phase-fraction in terms of WLR. The distributed phase fraction  is defined as  the percentage (or fraction) of water, oil, and gas in the total mixture fluids within the oil pipeline; the phase fractions are represented by the WLR and GVF. The WLR, or water cut, represents the volumetric fraction of water within the liquid component of the  multi-phase fluids and is a key parameter in the context of production optimization in the hydrocarbon industry, see Fig. \ref{fig:block-diagram} for block diagram on the use of DAS data for the industry.

Several state-of-the-art deep learning algorithms for DAS data, including ResNet~\cite{resnet}, SlowFastNet~\cite{slowfastnet}, DETR~\cite{detr}, and Perceiver~\cite{perceiver}, were used in this experiment. The ResNet model was built using the ResNet model for DAS data 
with 44-depth building blocks consisting of 42 bottleneck blocks. Each bottleneck block consist of [conv-bn-conv-bn-conv-bn-relu] layers structured sequentially per block. One-third of the bottleneck blocks use stride 1, while the rest use stride 2. The $\textbf{z}$ consist of 256 values, resulted from AdaptiveAvgPool2d of PyTorch. The SlowFastNet on the other hand was trained using transformed input DAS data by reshaping the input data to have four-dimensional data, e.g. sequence, spatial, temporal, and feature dimensions. The fast path of the SlowFastNet architecture samples every 2 items in the temporal dimensions, while the slow path samples every 16 items. Dropout and Batchnorm layers were included in the architecture with the $\textbf{z}$ consisting of 2056 values. 

The DETR model was trained using 3 encoders and 1 decoder block, with ResNet50 as the backbone block. 256 generated features are used as the outcome of the transformer layer, where they are being forwarded to the regression head. The last architecture is called Perceiver~\cite{perceiver}. The version of Perceiver with a depth of 3 and the number of latent dimensions of 128 was deployed for modelling the DAS data. For the Perceiver, the maximum frequency band was set to 10 and the number of bands was set to 6, following the default setting in the PyTorch version of Perceiver provided in Github.\footnote{https://github.com/lucidrains/perceiver-pytorch} Table \ref{baseline-results} shows the results of modelling the MPFF-DAS dataset using the deep learning algorithms.

\begin{table}[!t]
  \caption{Comparison of several state-of-the-art deep learning methods for DAS data on modelling WLR using the OOD DAS dataset.}
  \label{baseline-results}
  \centering
  \begin{tabular}{l cc}
    \toprule
    Model  & Train Error& Test Error \\
    \midrule
    ResNet                & $51.63$  & $60.26$ \\
    SlowFastNet           & $50.74$ & $68.18$  \\
    DETR                  & $31.54$ & \textbf{37.24}   \\
    Perceiver             & \textbf{29.46}& $39.06$   \\
    \bottomrule
  \end{tabular}
\end{table}

The data distribution of the target variables WLR is depicted in Figure \ref{fig:data-distribution}. Even though the DETR model was originally developed for object detection, it provides the lowest test error among the other models, see Table \ref{baseline-results}. Therefore, we treated the DETR model as the base model to study generalization techniques.

\begin{table}[!t]
    \caption{Test errors of several generalization techniques on estimating WLR using the OOD DAS dataset}
  \label{generalization-model}
  \centering
  \begin{tabular}[]{ll cc}
    \toprule
    
     Env. & Model &  Mean $\pm$ Std\\
    \midrule
    \multirow{5}{*}{Multi}
    & ERM & $36.82 \pm 0.93$  \\
    & IRM & $26.62 \pm 0.22$  \\
    & Rex & $36.21 \pm 1.04$  \\
    \cmidrule(lr){2-3}
    & Regr. Mixup ($\beta=1.1$) (ours)  &  $32.33 \pm 5.42$  \\
    & Regr. Mixup+IRM ($\beta=1.5$) (ours)  &  \textbf{26.53 $\pm$ 0.14}  \\
    \midrule
    
    \multirow{9}{*}{Single} 
    & ERM                      &  $35.02 \pm 5.55 $  \\
    
    & Ridge Regularization              &  $36.67 \pm 2.36$  \\
    & Lasso Regularization              &  $35.90 \pm 2.50$  \\

    & Input Mixup ($\alpha=1.0$)        &  $36.00 \pm 2.99$  \\
    & Manifold Mixup ($\alpha=1.0$)     &  $37.31 \pm 2.17$   \\
    & Manifold Mixup ($\alpha=2.0$)     &  $38.66 \pm 2.11$   \\
    
    \cmidrule(lr){2-3}
    
    & Regr. Mixup ($\beta=1.5$) (ours)     &  \textbf{26.74 $\pm$ 0.36}  \\
    \bottomrule
  \end{tabular}
\end{table}

Several generalization techniques were then deployed on the model, including Ridge Regularization, Lasso Regularization, IRM, Rex, Input Mixup, Manifold Mixup, and  OOD Regression Mixup. It should be noted that the DETR baseline model already includes  dropout, groupnorm, and batchnorm layers, thus the main objective is to better regularize the existing model on top of the existing regularizers. The REx and IRM were trained on multi-environment settings by slicing the MPFF-DAS training data in the time domain, providing three different environments. Each environment consists of around 40~minutes DAS recording; noise-perturbation is also included in each environment following experimental settings in~\cite{krueger2020out}. For completeness, we also included a normal ERM model,  OOD Regression Mixup with ERM, and OOD Regression Mixup with IRM on the multi-environment setting. Subsequently, the mixup algorithms and regularization norms were trained using single-environment data instead of the multi-environment settings. All the models were trained using learning rate \SI{1e-4} and SGD optimizer with momentum 0.9 using PyTorch library on a single node computer equipped with a NVIDIA P100 with 16 GB GPU memory, 90~GB~RAM, and 6 core Intel Xeon CPUs. We set the $\lambda$ to \SI{1e-4} for Ridge and Lasso Regularization, as well as our OOD Regression Mixup. We used the L1 loss function as our objective function and the accuracies were reported using the model that has the lowest training error in terms of MAE. With $b$ denotes the number of data points, the MAE is defined as:
\begin{align*}
    \text{MAE} = &   \sum_{a=1}^b \frac{|y_a -  \hat{y}_a|}{n}.
\end{align*}
The MAE is used because it proportionally measures how far off the prediction value is from the actual target value while providing a linear and easy interpretation of the phase-fraction error bars. To avoid statistical errors, we ran each experiment five times and the results are presented in Table \Ref{generalization-model}, in terms of the mean and standard deviation of MAE from multiple training sessions.

The experimental results show that the  OOD Regression Mixup results in smaller absolute errors of 8.28 and 4.49 compared to the baseline ERM model for single and multi-environment settings, respectively. It also validates our hypothesis that Input and Manifold mixup are least suitable for modelling the OOD regression dataset; our proposal provides significant improvement across all the settings in Table \Ref{generalization-model}. Interestingly, while the  OOD Regression Mixup with IRM provides the lowest error of 26.53 in the multi-environment setting, it is also more stable with a standard deviation of only $0.14$. For completeness, we also tested the algorithms by removing time windows in the data corresponding to transitions between different flow conditions when the flow had stopped (both WLR and GVF equal to zero). It shows that our OOD Regression Mixup~($\beta=1.5$) provides the lowest error of $29.42 \pm 0.19$, while the baseline ERM model provides an error of $30.95 \pm 2.76$.

\begin{table}[!t]
  \centering
    \caption{Comparison of several mixup-based techniques on Udacity dataset (in degree)}
  \begin{tabular}[t]{l c c}
    \toprule
    \multirow{1}{*}{Model}  & \multicolumn{1}{c}{Train Error} & \multicolumn{1}{c}{Test Error*} \\
    
     &  Mean $\pm$ Std &  Mean $\pm$ Std \\
    \midrule
       
    ERM                             & $3.39 \pm 0.57$ & $5.12 \pm 0.37$ \\
    Input Mixup ($\alpha=0.4$)      &  $8.29 \pm 0.45$ & $7.82 \pm 0.39$ \\
    Input Mixup ($\alpha=1.0$)      &  $8.19 \pm 0.99 $ & $7.61 \pm 0.41$ \\
    Manifold Mixup ($\alpha=1.0$)   &  $6.79 \pm 1.10$ & $6.58 \pm 0.81$ \\
    \midrule
    
    Regr. Mixup ($\beta=1.5$) (ours) & \textbf{3.07 $\pm$ 0.52} & \textbf{5.03 $\pm$ 0.32}  \\
    
    \bottomrule
  \end{tabular}
  \label{other-dataset-experiments}
\end{table}

\begin{table}[!t]
    \caption{Comparison of several mixup-based techniques on three different experimental settings of Rotation-MNIST dataset}
  \centering
  \begin{tabular}[b]{l ccc}
    \toprule
    \multirow{2}{*}{Model}  & \multicolumn{1}{c}{Slice-5} & \multicolumn{1}{c}{Slice-100} & \multicolumn{1}{c}{Slice-500}\\
    
                                    & Mean $\pm$ Std    & Mean $\pm$ Std    & Mean $\pm$ Std   \\
    \midrule
    ERM                             & $9.29 \pm 0.11$   & $9.37 \pm 0.42$   & $18.22 \pm 0.63$ \\
    Inp. Mix. ($\alpha=0.4$)      & $21.77 \pm 1.70$  & $24.50 \pm 4.12$  & $24.31 \pm 2.28$ \\
    Inp. Mix. ($\alpha=1.0$)      & $18.23 \pm 1.42$  & $18.60 \pm 1.59$  & $24.04 \pm 3.73$ \\
    Man. Mix. ($\alpha=1.0$)   & $16.71 \pm 1.00$  & $32.58 \pm 2.00$  & $23.87 \pm 1.91$ \\
    Man. Mix. ($\alpha=2.0$)   & $15.31 \pm 1.63$  & $15.00 \pm 0.94$  & $23.48 \pm 0.88$ \\

    \midrule
    
    Regr. Mix. ($\beta=1.1$) & \textbf{9.17 $\pm$ 0.11} & \textbf{9.14 $\pm$ 0.70} & \textbf{17.24 $\pm$ 0.64}\\
    
    \bottomrule
  \end{tabular}
  \label{mnist-dataset}
\end{table}

\subsection{Mixup on other regression datasets}\label{experiment-on-image-dataset}

For comparative evaluations, we also tested the proposed  OOD Regression Mixup on different datasets, including the Udacity and Rotation-MNIST datasets. The Udacity dataset consists of training and test data taken from videos of several real driving sessions recorded using three different cameras, e.g. left, right, and centre cameras. In this experiment, the training data were acquired only from the centre camera from two different driving scenes, namely HMB\_1 and HMB\_2. Because we do not have the label for the actual test data, we used the other driving sessions of training data, namely HMB\_4, HMB\_5, and HMB\_6 for the test data and retrieved the data from all three cameras. We used the 3D CNN model from~\cite{hara2018can} as the baseline ERM model and used 4-sequenced images as the input video to predict the steering angle (in degree) of the last image as the target prediction. We ran the experiments five times for each mixup algorithm and reported the results in terms of MAE in Table \Ref{other-dataset-experiments}. The results show that the decrease of mean MAE provided by the OOD Regression Mixup was considered minor compared to the baseline model of only $0.32$ and $0.09$ for the training and test data, respectively. However, it shows a significant improvement compared to the other existing mixup algorithms. For example, mixup regression leads reduced absolute errors when compared to Input mixup ($\alpha=1.0$) and Manifold mixup respectively of $5.12$ and $3.72$ for the training data, and $2.58$ and $1.55$ for the test data.

\begin{figure}[t]
  \centering
  \subfloat[Ablation experiments towards $\beta$ variable.\label{fig:training-distribution}]{%
       \includegraphics[width=0.99\linewidth]{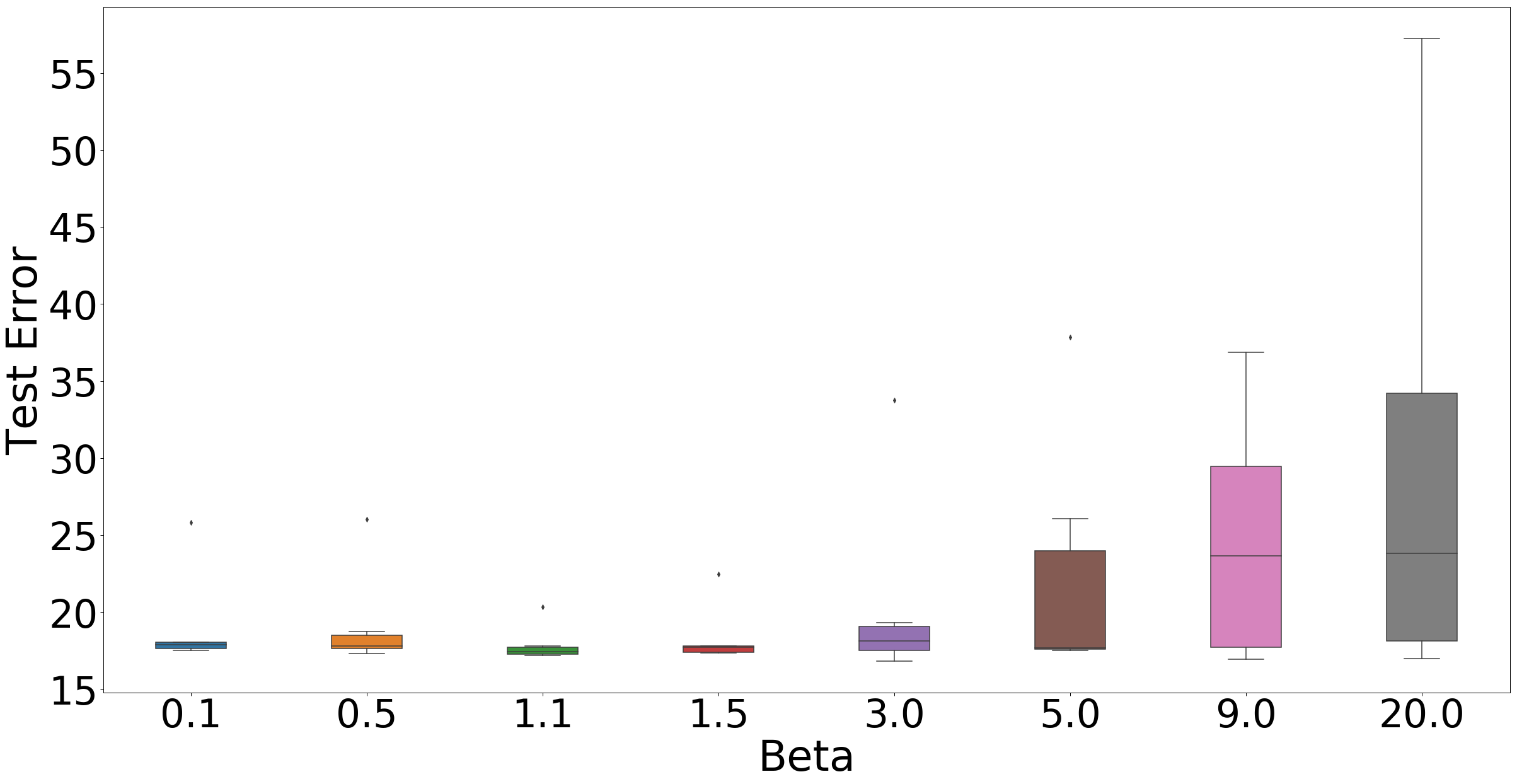}
       }
    \hfill
  \subfloat[Ablation experiments towards $\lambda$ variable.\label{fig:test-distribution}]{%
       \includegraphics[width=0.99\linewidth]{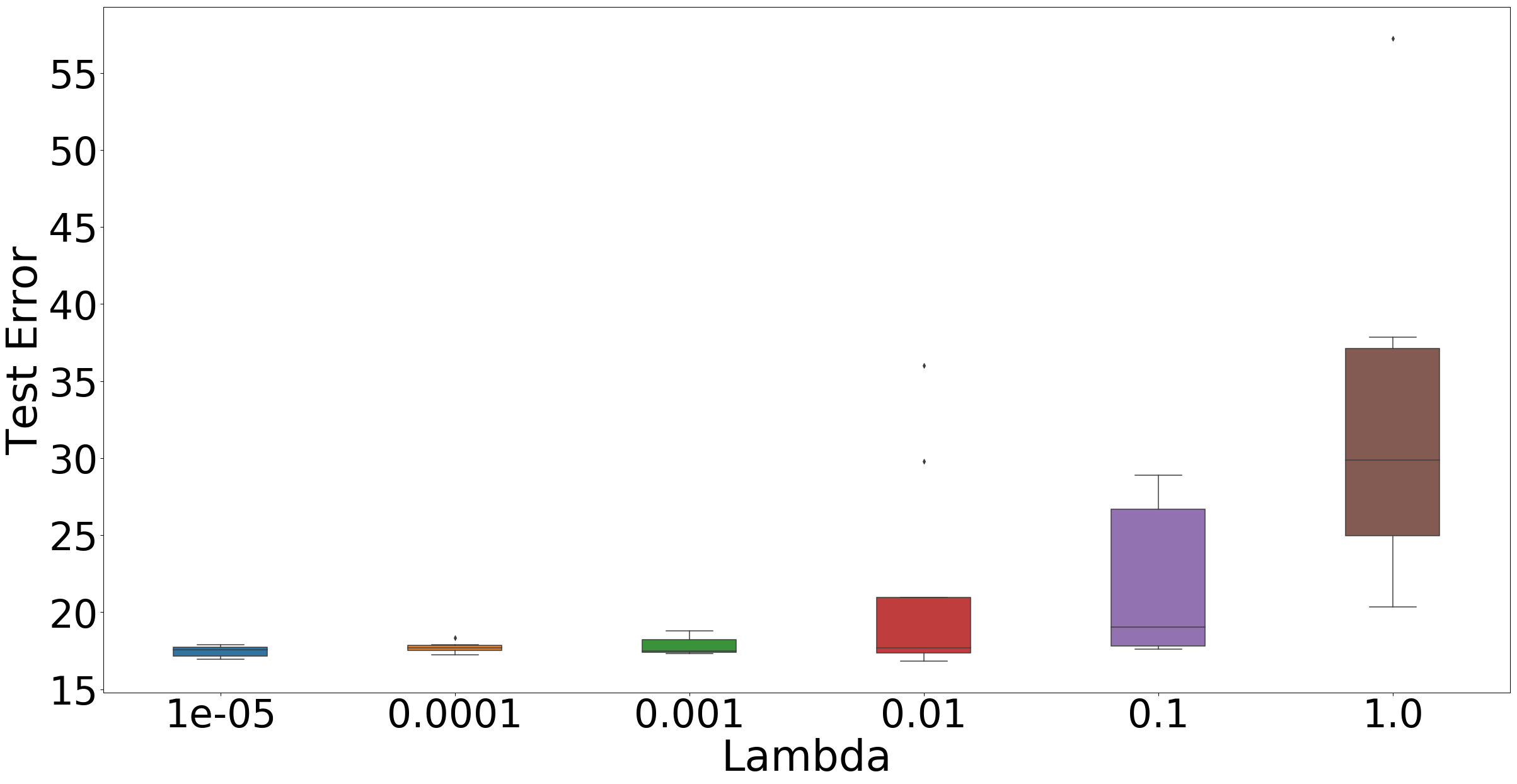}
       }
    
	\caption{Ablations experiments with the MNIST data to show the sensitivity of variable $\beta$ and $\lambda$, and the test error range resulted by changing those variables.
	}
	\label{fig:ablation-study-plot}
\end{figure}

In contrast to the Udacity dataset, the Rotation-MNIST is a toy data, created by rotating MNIST data between 0 and 180 degrees with 10 different samples on each degree, resulting in 1800 unique target values (rotation angles). Each unique value is represented using 500 different images from 10 different MNIST classes. 
A small CNN, called Resnet-18, with an initial learning rate of \SI{1e-2} was used as the base model for predicting the rotation angle for each input image. We ran three contrasting experimental settings representing the OOD problems on the regression dataset based on the existence of the target variables in the training data. We removed half of the training data using a sequentially-sliced removal method based on the target values. First, the training data were sorted based on all the 1800 target values, then, 5 different target values were removed sequentially for every 5 other different target values (representing a range of 0.5-degree removal per step); the setting is called slice-5. We performed a similar slicing method by using 100 (slice-100) and 500 (slice-500) different target values to represent a more challenging OOD problem representing a range of 10 and 50-degree removal per step, respectively. The test data, on the other hand, was generated using similar process from building the training set, but the whole Rotation-MNIST test data without any removal process (around 18~000~000 data points) were used to evaluate the generalization performances. We ran the experiments five times and reported the mean and standard deviation (std) in terms of MAE for each setting. 

The results are presented in Table \Ref{mnist-dataset} and show that the proposal works far superior to other mixup algorithms and is comparable with normal ERM. Subsequently, with a more challenging problem (OOD shift) in slice-500, the proposal works better than ERM, showing that our mixup can work better on the OOD dataset while proposing a good result for the normal regression dataset.

Fig. \ref{fig:test-mnist-distribution} shows that the input mixup (blue dotted line) works better than our proposal in a small range of rotation degrees but it underfits the data for most of the distribution, making it an unreliable form of OOD generalization technique. On the other hand, our OOD Regression Mixup not only works well along with the training distribution and performs better than the empirical model within this distribution, but it also works superiors on the OOD data where the empirical data are non-existent. These results confirm our hypothesis further, that by leveraging the linearity of two similar data points and their manifold hidden variables, our proposal can provide a model that is not only fitting well with the existing data but can only generalize on the unseen OOD data; a characteristic that does not exist in the ERM model and ignored by the existing mixup-based techniques in the benchmark. 

\subsection{Computation cost}\label{computation-cost}

The computation cost from our proposal comes from calculating the distance kernel (Eq. \ref{eq:contrast_sensitive_kernel}) and proportional distance of~$z$ (Eq. \ref{eq:proportional_z}); they have a linear complexity implemented using simple mathematical formulations. For example, on modelling the MPFF-DAS dataset on inference/test mode, 11.64 and 10.93 samples per second can be proceeded by the ERM and OOD Regression Mixup models, respectively. It is worth noting that the network speed, hard-drive utilization, and CPU load are varied during the processing time, thus making the small difference in inference time between the two models can be ignored. 

In this study, each sample contains 5 seconds of information, thus processing a few samples per second for the MPFF objective can provide a real-time application. Moreover, as the additional computation cost from our algorithm is negligible on modern hardware, the proposal can be implemented to better regularize the existing real-time deep learning algorithms, including Fast RCNN, YOLO, and EfficientNet on embedded devices.     

\subsection{Ablation study}\label{ablation-study}

\begin{table}

  \centering
  \caption{Comparison of different parameter settings on the  OOD Regression Mixup for modelling the Rotation-MNIST dataset}
 \begin{tabular}[t]{l | ccc ccc}
   \toprule
               \boldsymbol{$\beta$}~\textbackslash~\boldsymbol {$\lambda$}    & $1e$-05  & $1e$-04& 
                     $0.001$ & 
                      $0.01$ & 
                       $0.10$& 
                        $1.00$ \\

    \midrule
$0.10$ & 17.96 & 18.07 & 18.01 & 17.66 & 17.79 & 22.77 \\
$0.50$ & 17.36 & 17.33 & 17.50 & 17.42 & 19.52 & 20.05 \\
\midrule
$1.10$ & \textbf{16.84} & 17.24 & 17.54 & 17.19 & 17.08 & 19.84 \\

$1.50$ & 17.43 & 17.83 & 17.37 & 17.01 & 17.81 & 19.81 \\
\midrule
$3.00$ & 17.84 & 18.44 & 17.50 & 16.88 & 18.11 & 22.85 \\
$5.00$ & 17.49 & 17.69 & 17.34 & \textbf{16.75} & 23.62 & 34.92 \\
$9.00$ & 16.99 & 17.22 & 18.04 & 21.18 & 25.03 & 31.56 \\
$20.00$ & 16.95 & 17.63 & 18.77 & 19.76 & 25.96 & 49.05 \\

    \bottomrule
 \end{tabular}
  \label{ablation-results}
\end{table}

We turn to the ablation study to see the effect of each parameter of our proposal on the OOD setting of the rotation-MNIST dataset. The proposal uses $\beta$ and $\lambda$ as additional variables, therefore, we used several different $\beta$ and $\lambda$ on the ablation experiments, and the results are presented in Table \Ref{ablation-results}. The results show that by providing a relatively low $\lambda$, the value of $\beta$ can relatively be ignored on achieving a lower regression error. The same results can also be achieved by providing a low $\beta$ and ignoring the values of  $\lambda$, depicted in Fig. \ref{fig:ablation-study-plot}a and Fig. \ref{fig:ablation-study-plot}b. These interesting results simplify the search space for finding the optimal values for both $\beta$ and $\lambda$. By keeping the low values of $\lambda$, therefore $\lambda$ $\leq$ \SI{1e-4} and $\beta$ $\sim$ 1.0, the proposal could achieve a good generalization performance, as shown empirically in Table \ref{ablation-results} and Fig. \ref{fig:ablation-study-plot}.

The ablation study also reveals that setting up the $\lambda$ values high makes the generalization results unreliable. This phenomenon is expected, because a high $\lambda$ enforce the linearization within the feature spaces towards rank-1, limiting the capability of the underlying model to represent the complexity of the input data. To secure the solution space, we suggest the $\lambda$ is set between 0.01 and \SI{1e-6}, we argue that setting $\lambda$ lower than \SI{1e-6} will make the model perform very similar to ERM because a very low $\lambda$ will make the Mixup Regularization Loss (in Eq.~\ref{eq:regression_mixup}) equals to the ERM Loss (Eq.~\ref{eq:erm}).

\subsection{Limitations}\label{limitation}
The main limitation of our work is on choosing the optimized values of $\beta$ and $\lambda$ because we do not have a good theoretical basis and optimization strategy to find the optimized version for both parameters. However, based on the experimental study, setting $\beta=1.1$ and $\lambda=$~\SI{1e-4} works fine for many cases. Lastly, because we aim to minimize the proportional distances on $\textbf{z}$ while considering similarity values on $y$, the algorithm might only work on smooth regression datasets and might not work on high volatility regression data, such as the stock market datasets. This limitation, unfortunately, makes the proposal have limited use outside the smooth regression datasets.

The future studies of our work include: firstly, to find a way to automate the finding of the optimum values of the parameters in our algorithm, therefore the algorithm can be parameter-free and is easier to be adopted  within the community. The ablation study shows a possible correlation between $\lambda$ and $\beta$, which suggest that optimization techniques, such as gradient-based search and genetic algorithms, can be used for optimizing the two dependent variables. Secondly, as our research aims to optimize hydrocarbon production, the future works must emphasize the use of the MPFF-DAS dataset. This dataset is unique and can help provide a game-changing functionality for monitoring hydrocarbon production in the oil and gas industries. 
Moreover, as the research in acoustic signal progress, our proposal can also be a complement to other research in the field, including \cite{faultdiagnosticusingacoustic} fault diagnostic with machine learning. Acoustic signals in low frequencies for fault diagnostic have intrinsic linearization features, e.g. smoother changes along the time dimension during the acoustic events. We argue that this characteristic make the proposed generalization algorithm more compatible with the research objectives.

\section{Concluding Remarks}\label{conclusions}

We have presented the OOD Regression Mixup, a simple and intuitive cost function based on the properties of existing mixup algorithms and regularization norms to provide a better way to model regression data, especially for OOD datasets. We also have provided a spatio-temporal  dataset, MPFF-DAS. This dataset is unique and provides a stepping stone to understand the wide range of acoustic signatures from multi-phase fluids captured by the fiber optic cable. 

Throughout an extensive evaluation, we have shown that our OOD Regression Mixup algorithm provides a much lower regression error compared to the existing ERM models and other mixup algorithms on several regression-based datasets, including Udacity, Rotation-MNIST, and the spatio-temporal MPFF-DAS dataset. In our experiments, we found out that the OOD Regression Mixup works well with several different types of neural network architectures, including the ResNet, 3D CNN, and attention based-model. The algorithm also provides a robust generalization model as indicated by  small training session stochasticity based on the values of standard deviations from multiple experiments. Moreover, using a relatively small $\lambda$, we have shown that the experimental evidence supports our hypothesis that some degree of linearity exists within the regression data, and taking advantage of this could provide a better generalization model for the corresponding dataset, especially for the OOD datasets. 


\bibliographystyle{IEEEtran}
\bibliography{IEEEtran}

%



\vfill



\end{document}